\documentclass[conference]{IEEEtran}
\usepackage{amscd}
\usepackage{supertabular}
\usepackage{amssymb}
\usepackage{rotating}
\usepackage{amsmath}
\usepackage{graphicx,algorithm,algorithmic}
\usepackage{subfigure}
\usepackage{epsfig}
\usepackage{theorem}
\usepackage{lscape}
\usepackage{float, flushend}
\usepackage{booktabs}
\usepackage{fancybox}
\usepackage{txfonts}
\usepackage{mathrsfs}
\usepackage[usenames]{color}
\usepackage{comment}
\usepackage{fancyhdr}
\usepackage{fancyheadings}

\newcommand{\bs}[1]{\boldsymbol{#1}}
\newcommand{\beqn}{\begin{equation}} 
\newcommand{\eeqn}{\end{equation}} 

\begin{document}
\title{Radon Features and Barcodes\\ for Medical Image Retrieval via SVM}

\author{\IEEEauthorblockN{Shujin Zhu}
\IEEEauthorblockA{School of Electronic \& Optical Engineering\\
Nanjing University of Science \& Technology\\
Jiangsu, China 210094}
\and
\IEEEauthorblockN{H.R. Tizhoosh}
\IEEEauthorblockA{KIMIA Lab\\University of Waterloo, Canada\\
Email: tizhoosh@uwaterloo.ca}
}

\maketitle

%\thispagestyle{fancy}
%\lhead{} 
%\chead{} 
%\rhead{} 
%\lfoot{} 
%\cfoot{} 
%\rfoot{\thepage} 
%\renewcommand{\headrulewidth}{0pt} 
%\renewcommand{\footrulewidth}{0pt} 
%\pagestyle{fancy} 
%\rfoot{\thepage} 
\begin{abstract}
For more than two decades, research has been performed on content-based image retrieval (CBIR). By combining Radon projections and the support vector machines (SVM), a content-based medical image retrieval method is presented in this work. The proposed approach employs the normalized Radon projections with corresponding image category labels to build an SVM classifier, and the Radon barcode database which encodes every image in a binary format is also generated simultaneously to tag all images. To retrieve similar images when a query image is given, Radon projections and the barcode of the query image are generated. Subsequently, the k-nearest neighbor search method is applied to find the images with minimum Hamming distance of the Radon barcode within the same class predicted by the trained SVM classifier that uses Radon features. The performance of the proposed method is validated by using the IRMA 2009 dataset with 14,410 x-ray images in 57 categories. The results demonstrate that our method has the capacity to retrieve similar responses for the correctly identified query image and even for those mistakenly classified by SVM. The approach further is very fast and has low memory requirement.
\end{abstract}

\IEEEpeerreviewmaketitle

\section{Introduction} 
Content-based image retrieval (CBIR) aims at searching for similar images in a database when a user provides a query (input) image. CBIR algorithms are supposed to work with measures derived from the images themselves instead of accompanied text or metadata \cite{Azevedo2013}. In recent years, CBIR has been investigated for different fields such as education, electric commence, Internet and bio-medical area. 

CBIR systems rely on analysis of the content of the query image and efficient methods to describe the visual features extracted from the image. Early CBIR systems were based on the descriptions of visual characteristics, such as color, texture, and shape, and advanced features representing the vital regions with significant details \cite{shyu2003image,sun2004hierarchical,ko2004svm}, or in more recent works the semantic content \cite{li2010object,kuo2012unsupervised,jin2015heterogeneous}. In the medical field, CBIR systems have been examined with the hope that they can assist clinicians in diagnostic fields by retrieving similar cases. 

In \cite{verma2015center}, authors extract the center-symmetric local binary pattern (LBP) from image and compute co-occurrence of pixel pairs in local pattern map using gray-level co-occurrence matrix, which is assumed to be more robust than frequency of patterns (histogram). Dubey et al. \cite{dubey2015local} propose a new feature descriptor named local diagonal extrema pattern (LDEP) for CT image retrieval. By using the first order local diagonal derivatives, the values and indexes of the local diagonal extremes are obtained, and the descriptor is generated on the basis of the indexes and the relationship between the center pixel and local diagonal extremes. Greenspan et al. present a localized statistical framework based on Gaussian mixture modelling and Kullback-Leibler (KL) matching for medical image retrieval. It combines a continuous, probabilistic and region-based image representation scheme, along with an information-theoretic image matching measure to categorize and retrieve x-ray images \cite{greenspan2007medical}. Camlica et al. used two methods to retrieve x-ray images: an autoencoder to reduce the size of the image for feature extraction \cite{camlica2015autoencoding}, and a SVM classifier trained with LBP features derived from saliency-based image regions. \cite{camlica2015SVM}. 

It is worth noting that most current medical CBIR systems are limited to the images with specific modality, organ, or diagnostic studies, and usually cannot be applied directly to other medical applications \cite{lehmann2004content}. Moreover, although CBIR can theoretically be applied to several medical cases, the existence of ``semantic gap'' between the extracted image features on one side and the doctors' interpretation of the images on the other side limits its application. This is because generic features of images may not be suitable for the analysis of medical image similarity of a specific modality \cite{Azevedo2013}. 

Inspired by the barcodes embedded in many products and attached to many services we encounter on a daily basis, a novel Radon barcode for medical image retrieval system was proposed in 2015 \cite{tizhoosh2015barcode}. The Radon barcode is a binary code generated based on Radon Transform with selected projection angles and projection binarization operation that can tag/annotate a medical image or its regions of interest. Depending on the Radon barcode, large image archives can be efficiently searched to find matches via Hamming distance, e.g., kNN search,  or hashing-based methods, e.g., locality sensitive hashing. 

In this paper, IRMA benchmarking dataset containing 14,410 x-ray images are used to test an approach based on SVM and Radon transform. By using Radon transform, the medical images in database are described by a set of features to train support vector machines using corresponding image labels representing their categories or classes. Radon barcodes of all images in database are also generated for the actual retrieval. When new query (input) images are encountered, SVM classifier first assigns them a class by classifying their Radon projections. Within the class determined by SVM, the most similar images are then retrieved by using the $k$-nearest neighbor (KNN) search method that compares Radon barcode of query image with the barcodes of all other images already indexed.

This paper is organized as follows: Section \ref{sec:bkg} provides a brief overview of SVM, Radon transform and Radon barcodes. The proposed approach is described in \ref{sec:prop}. The experiments and results will be reported in section \ref{sec:exp}. The paper will be concluded in section \ref{sec:conc}. 

\section{Brief Review of SVM and Radon Barcodes}
\label{sec:bkg}
In this section we briefly review SVM, Radon transform and Radon barcodes. 

\subsection {Support Vector Machines (SVM)}
Given a set of separable training samples which belong to two different classes, a linear SVM constructs a hyperplane which contains the largest number of samples of the same class on the same side, while achieving the largest distance to the nearest training-data point of any class by the hyperplane \cite{Vapnik1995,cusano2003image}. The general form of linear SVM can be denoted as
\beqn
y_i(\texttt{w}\cdot \texttt{x}_i+b)\geq 1, \forall i\in \left\{1,\cdots,N\right\},
\eeqn
where $b$ is a bias, $\texttt{x}_i$ is the training samples $(i=1,\cdots, N)$ and $y_i=\left\{-1, 1\right\}$ is the class label where the sample $\texttt{x}_i$ belongs. The distance from the closest point to the hyperplane is $1/\texttt{w}$. Finding the maximal distance to the closet point, which is called optimal separating hyperplane (OSH), amounts to minimizing $\texttt{w}$ and the objective functions is
\beqn
min \phi(\texttt{w})=\frac{1}{2}\left\|\texttt{w}\right\|^2 
\eeqn
subject to
\beqn
y_i(\texttt{w} \cdot \texttt{x}_i+b)\geq 1, \forall i\in \left\{1,\cdots,N\right\}.
\eeqn
When the training set is not linearly separable, a set of ``slack variables'' $e$ and a penalization \textsl{C} for misclassified cases are introduced and the objective function can be written as \cite{zhang2001support}
\beqn
min\left(\frac{1}{2}\left\|\texttt{w}\right\|^2+\textit{C}\sum^{\textit{N}}_{\textit{i}=1}\texttt{e}_\textit{i}\right)
\label{objective_function_unlinear}
\eeqn
under the constraints
\beqn
y_i(\texttt{w} \cdot \texttt{x}_i+b)\geq 1-\texttt{e}_\textit{i}, \texttt{e}_\textit{i}\geq 0, \forall i\in \left\{1,\cdots,N\right\}.
\eeqn

Given a new sample, the label $g$ can be predicted without explicitly computing $\texttt{w}$:

\beqn
g(\texttt{x})= sgn\left( b+\sum^{N}_{i=1} \alpha_i y_i \left(\texttt{x}_\textit{i}\cdot\texttt{x}\right)\right),
\eeqn
where $\alpha$ is the Lagrange Multiplier. For multiple classes, the multi-class classifier can be obtained by combining the result of each classifier that denotes a discrimination function $f_i$ that delivers positive values to the samples in the same class and negative values for the ones belong to different class. The discrimination function can be written as 
\beqn
f_i(\bs{x})=\frac{\texttt{w}_i\cdot \texttt{x}+b_i}{\left\|\texttt{w}_i\right\|} ,
\eeqn 
where $\texttt{w}_i$ and $b_i$ are the normal in and bias of the hyperplane in the $i$-th SVM, respectively. For the input new case $\texttt{x}$, its label can be predicted by multi-class SVM classifier as
\beqn
c(\texttt{x})= \operatorname*{arg\,max}_{\textit{i}\in\left\{1,\ldots,K\right\}}f_i\left(\texttt{x}\right),
\eeqn
where $K$ is the number of classes.

\subsection {Radon Transform}
The Radon transform is the integral transform which computes the projection of the image along various directions. The Radon transformation was introduced by J. Radon in 1917 \cite{radon1986determination} as a way of reconstructing a function with the values of its projections in $\textbf{R}^n$ space \cite{surender2012representation}. It is widely applicable to many areas such as computed axial tomography, reflection seismology, barcode scanners and computer vision. The Radon transform of a two-dimensional image $f(x,y)$ can be defined as its line integral along a line inclined at an angle $\theta$ and at a distance $\rho$ from the origin \cite{seo2004robust}:
\beqn
R(\rho,\theta)=\int^{\infty}_{-\infty} \int^{\infty}_{-\infty} f(x,y) \delta(x cos\theta+y sin\theta-\rho) dx dy,
\eeqn
where $\delta(\cdot)$ is the Dirac Delta function which is non-zero only on $s$ axis and its integral, and $-\infty<\rho<\infty$,$0\leq\theta <\pi$. The Radon transform has the capacity to accentuate straight-line features from an image by integrating the image intensity over the straight line to a single point \cite{rey1990application}. 

\subsection{Radon Barcodes}
%An image $I$ is a 2D function $f(x,y)$. One can project $f(x,y)$ along a number of projection angles using Radon transform. The projection is basically the sum (integral) of $f(x,y)$ values along lines constituted by each angle $\theta$. The projection creates a new image $R(\rho,\theta)$ with $\rho = x \cos \theta + y \sin \theta$. Hence, using the Dirac delta function $\delta(\cdot)$ the Radon transform can be written as 
%\begin{equation}
%R(\rho,\theta) = \int\limits_{-\infty}^{+\infty} \int\limits_{-\infty}^{+\infty} f(x,y) \delta(\rho-x\cos \theta-y\sin\theta) dx dy.
%\end{equation}

\begin{figure}[htbp]
\begin{center}
\includegraphics[width=0.90\columnwidth]{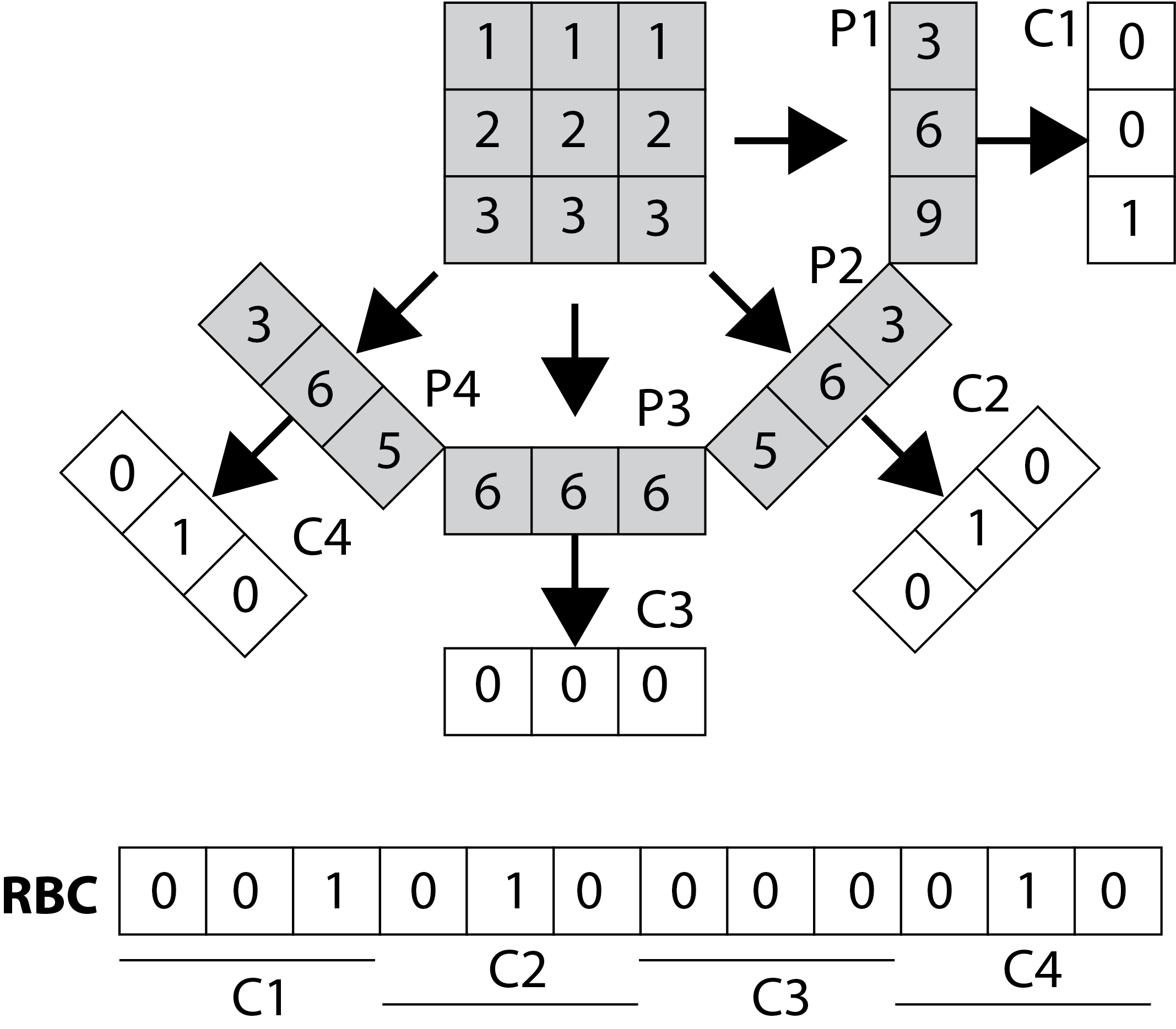}
\caption{Radon Barcode (RBC) \cite{tizhoosh2015barcode} -- Projections P1, P2, P3, and P4 are thresholded to generate code fragments C1, C2, C3 and C4. The concatenation of all code fragments delivers the barcode \textbf{RBC}. }
\label{fig:RBC}
\end{center}
\end{figure}

As proposed in \cite{tizhoosh2015barcode}, one can threshold all projection values for individual angles based on a ``local'' threshold for that angle \cite{Tizhoosh2016}. Subsequently, one can assemble a barcode of all thresholded projections as depicted in Figure \ref{fig:RBC} \cite{tizhoosh2015barcode}. A simple way for thresholding the projections is to calculate a typical value via median operator applied on all non-zero values of each projection. Algorithm \ref{alg:Radon} describes how \textbf{Radon barcodes (RBC)} are generated \footnote{Matlab code available online: http://tizhoosh.uwaterloo.ca/}. In order to receive same-length barcodes \emph{Normalize$(I)$} resizes all images into $R_N\times C_N$ images (i.e. $R_N= C_N=2^n,n\in \mathbb{N}^+$).

% Radon Barcode -------------------------
\begin{algorithm}[htbp]
\caption{Radon Barcode (RBC) Generation \cite{tizhoosh2015barcode}}
\begin{algorithmic}[1]
\label{alg:Radon}
\STATE Initialize Radon Barcode $\mathbf{r} \leftarrow \emptyset$ 
\STATE Initialize angle $\theta \leftarrow 0$ and $R_N=C_N\leftarrow 32$
\STATE Normalize the input image $\bar{I} = \textrm{Normalize}(I,R_N,C_N)$ 
\STATE Set the number of projection angles, e.g. $n_p \leftarrow 8$
\WHILE{$\theta < 180$}
\STATE Get all projections $\mathbf{p}$ for $\theta$
\STATE Find typical value $T_\textrm{typical}\leftarrow\textrm{median}_i (\mathbf{p}_i)|_{\mathbf{p}_i \neq 0}$
\STATE Binarize projections: $\mathbf{b} \leftarrow \mathbf{p} \geq T_\textrm{typical}$ 
\STATE Append the new row $\mathbf{r} \leftarrow \textrm{append}(\mathbf{r},\mathbf{b} )$ 
\STATE $\theta \leftarrow \theta + \frac{180}{n_p}$
\ENDWHILE
\STATE Return $\mathbf{r}$
\end{algorithmic}
\end{algorithm}

\begin{figure*}[htbp]
\begin{center}
\includegraphics[width=\textwidth]{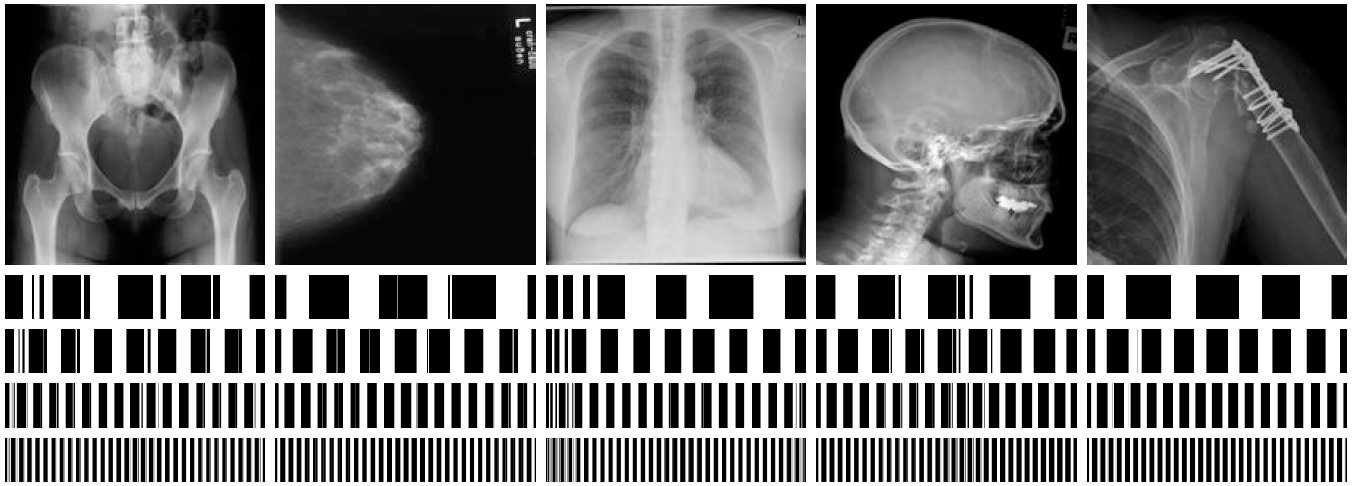}
\caption{Five randomly selected images from IRMA dataset and their Radon barcodes with (from top to bottom) 4, 8, 16 and 32 projections.}
\label{fig:sampleBarcodes}
\end{center}
\end{figure*}

Figure \ref{fig:sampleBarcodes} show some sample images from IRMA dataset and their Radon barcodes with 4, 8, 16 and 32 projections applied on input images resized to $64\times 64$.

\section{The Proposed Method}
\label{sec:prop}
In this section, the proposed image retrieval procedure is described. Basically, the proposed method is composed of two stages: classification and retrieval. During a training stage, each image in the dataset is normalized to the same dimension and the Radon transform is used to extract the Radon features. SVM classifier is then trained by using the extracted Radon features together with their corresponding class labels. The Radon barcode of training dataset is also generated in the training stage to index all images within each class. For retrieval stage, Radon barcodes are obtained and $k$-nearest neighbor search is employed to retrieve similar cases.

\textbf{Radon Features --} To extract Radon features, the dimensions of every image in the training dataset are required to be normalized to a fixed size in order to same-length features and barcodes (generally a down-sampling task to reduce the computational burden). In this work, we experiment with the preferable size $(R_N \times C_N)$ to be $16 \times 16$, $32 \times 32$ or $64 \times 64$. Then the Radon transform is performed with a preset number of projection angles to generate the corresponding Radon projections. The Radon features are then normalized within $[0,1]$. The normalized Radon features are then used for both training the SVM classifier and for subsequent Radon barcode generation.

\textbf{Training Stage --} Combining with the label that defines the image class, the Radon features are used to train C-support vector multi-class SVM\footnote{https://www.csie.ntu.edu.tw/$\sim$cjlin/libsvm/}. The multi-class SVM used in this procedure implements the ``one-against-one'' approach \cite{knerr1990single}, which has been shown to be a very robust and accurate method by Hsu and Lin \cite{hsu2002comparison}. The SVM kernel is set to Radial Basis Function. The setting of other parameters is empirical. In the training stage, the Radon projections are binarized to generate a Radon barcode. Algorithm \ref{alg:training} shows the generic overview of the training stage with IRMA dataset. 

\begin{algorithm}[htbp]
\caption{Training Stage}
\begin{algorithmic}[1]
\label{alg:training}
\STATE Initialize angle $\theta$ and image size $R_N=C_N\leftarrow N$
\WHILE{$i <$ training samples}
\STATE Normalize the image $\bar{I}_i=Normalize(I_i,R_N,C_N)$
\STATE Radon transformation $R_i=Radon(\bar{I}_i)$
\STATE Binarization projections:$b_i\leftarrow R_i\geq T_{threshold} $
\STATE Generate radon barcode $B_i=Barcode(b_i)$
\ENDWHILE
\STATE Get the label $\textbf{L}$ 
\STATE Build SVM classifier \textbf{M}$=$trainSVM $(\textbf{b},\textbf{L})$
\end{algorithmic}
\end{algorithm}
\textbf{Retrieval Stage --} In the retrieval stage, the query image is normalized by the same approach as applied to training samples, and is processed to generate its corresponding Radon projection and Radon barcode. The Radon projection is then used to assign a class to the image by the multi-class SVM classifier. According to the classified image label, the images within the same category are selected from the database, and the Radon barcode generated from the query image is compared with all images within that class by using the $k$-nearest neighbor searching method. Eventually, the most similar images in the same class can be indexed out. Algorithm \ref{alg:retrieval} shows the overview of the retrieval stage for query image.

\begin{algorithm}[htbp]
\caption{Retrieval Stage}
\begin{algorithmic}[1]
\label{alg:retrieval}
\STATE Query Initialization: $\bar{I}_q = Normalize (I_q,R_N,C_N)$
\STATE Generate Radon Projection: $R_q = Radon(\bar{I}_q)$
\STATE Predict image class: $label_p = predictSVM(R_q,\textbf{M})$
\STATE Generate query RBC: $B_q = Barcode(R_q)$
\STATE Obtain Radon barcode: $\textbf{B}_p\leftarrow=B_i|_{i=label_p}$
\STATE Set the desired number $k$ of retrieved images
\STATE Search the top $k$ images:$\textbf{I}_k = KNN(\textbf{B}_p,B_q,k)$
\STATE Return retrieved images $\textbf{I}_k$
\end{algorithmic}
\end{algorithm}

\begin{figure*}[tb]
\begin{center}
\includegraphics[width=0.8\textwidth]{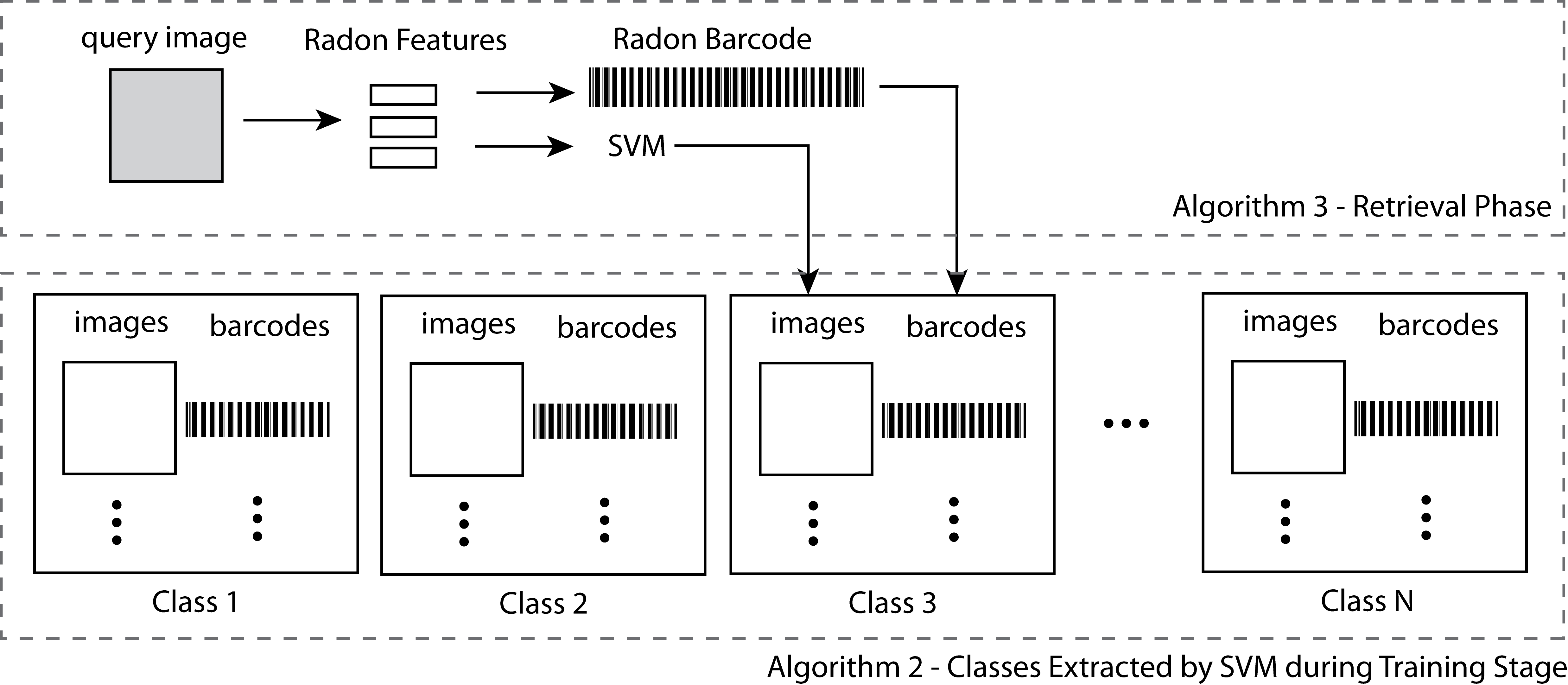}
\caption{Training and Retrieval Stages according to Algorithms \ref{alg:training} and \ref{alg:retrieval}. }
\label{fig:overallscheme}
\end{center}
\end{figure*}

Figure \ref{fig:overallscheme} illustrates the overall proposed scheme for classification (Algorithm \ref{alg:training}) and retrieval (Algorithm \ref{alg:retrieval}).

\section{Experiment and Discussions}
\label{sec:exp}
In this section we describe the details of the benchmarking data and the error calculation to report the experiments for both classification using SVM with Radon features and retrieval using Radon barcodes. 
\subsection{Image Test Data }
The IRMA database with a collection of 14,410 x-ray images taken arbitrary form medical routine is used for training and testing \cite{tommasi2010overview,Lehmann2003,Lehmann2006}. The images represent different ages, genders, view positions and pathologies, which have been classified in 57 categories by scanning body position. A total of 12,677 x-ray images are collected in the training dataset and the other 1,733 images are considered as testing data. For each image in the IRMA dataset, a string of 13 characters with 4 axes named IRMA code is annotated (TTTT-DDD-AAA-BBB) which describes the image modality, body orientations, body region examined and its biological system. 

\subsection{Error Measurements}
Based on the IRMA code, the error score computation approach which evaluates the retrieval performance is defined as \cite{tommasi2010overview}
\beqn
\sum^{I}_{i=1} \frac{1}{b_i} \frac{1}{i} \delta(l_i, \hat{l_i}),
\eeqn
where $b_i$ is the number of the possible labels for position $i$ and $\delta$ is the Delta function delivering $0$ if $l_i=\hat{l_i}$, otherwise $1$. 

\subsection{Classification Performance}
In the IRMA database, some images both in training and testing datasets have missing labels. Thus, the data without labels are ignored. Then, the remaining $12,631$ images in training data ($42$ non-labeled training images) and $1,639$ images for testing ($94$ non-labeled testing images) are used in our experiments. 

In this paper, we choose $k(x,y)= e^{-\rho\left\| x-y \right\|^{2}_{2}}$ as the kernel in the experiments and the parameter $\rho$ and the penalization coefficient $C$ of Eq.(\ref{objective_function_unlinear}) are set empirically to $0.0359$ and 16, respectively. To evaluate the performance of the classifier, the classification accuracy of SVM is defined as
\beqn
\textrm{Accuracy}=\frac{| \textrm{correctly predicted data} |}{| \textrm{total testing data} |} \times 100\%.
\eeqn

The results of classification applied on IRMA dataset with various image sizes and different numbers of projection angles are shown in Table \ref{tab:SVMresults}. The results suggest that the classification accuracy is improved with the increase in the number of Radon projection angles, a result that is consistent with general knowledge in image reconstruction (the more projections the more accurate the image reconstruction).

\begin{table}[!t]
\caption{Classification accuracy of SVM applied on IRMA dataset} 
\centering 
\begin{tabular}{cccc} 
\hline
& & Normalized Size \\ \hline
Num. of Projections & $16 \times 16 $ & $32 \times 32$ & $64 \times 64$ \\ \hline 
8 & 51.86\% & 54.91\% & 55.70\% \\ 
16 & 53.38\% & 56.13\% & 56.07\% \\ 
32 & 55.27\% & 56.92\% & 55.33\% \\ \hline
\end{tabular}
\label{tab:SVMresults}
\end{table}

\begin{figure}[!t]
\centering
\includegraphics[width=.5\textwidth]{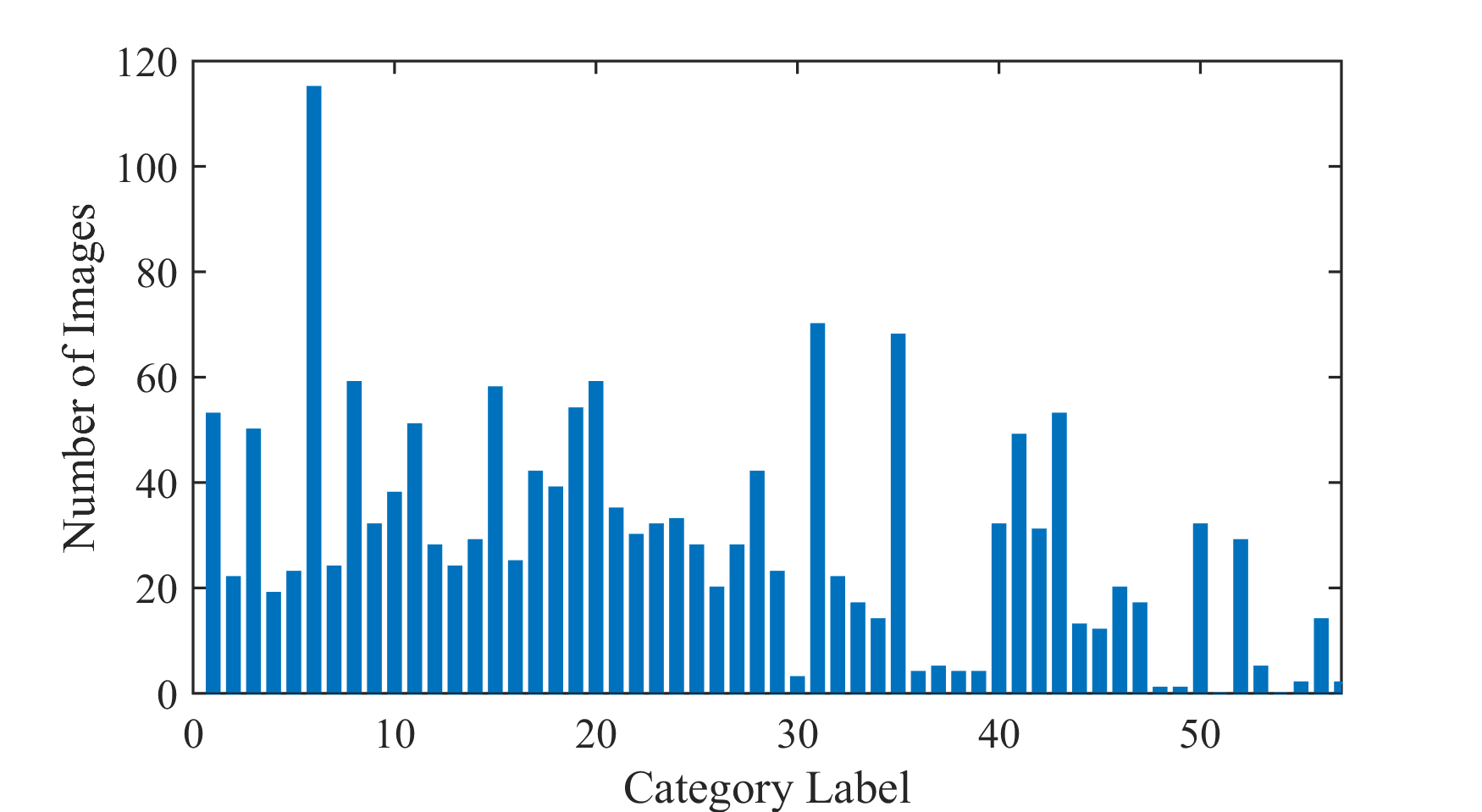}
\caption{Number of images in each class}
\label{fig:number_of_images_each_class}
\end{figure}

\begin{figure}[!t]
\centering
\includegraphics[width=.5\textwidth]{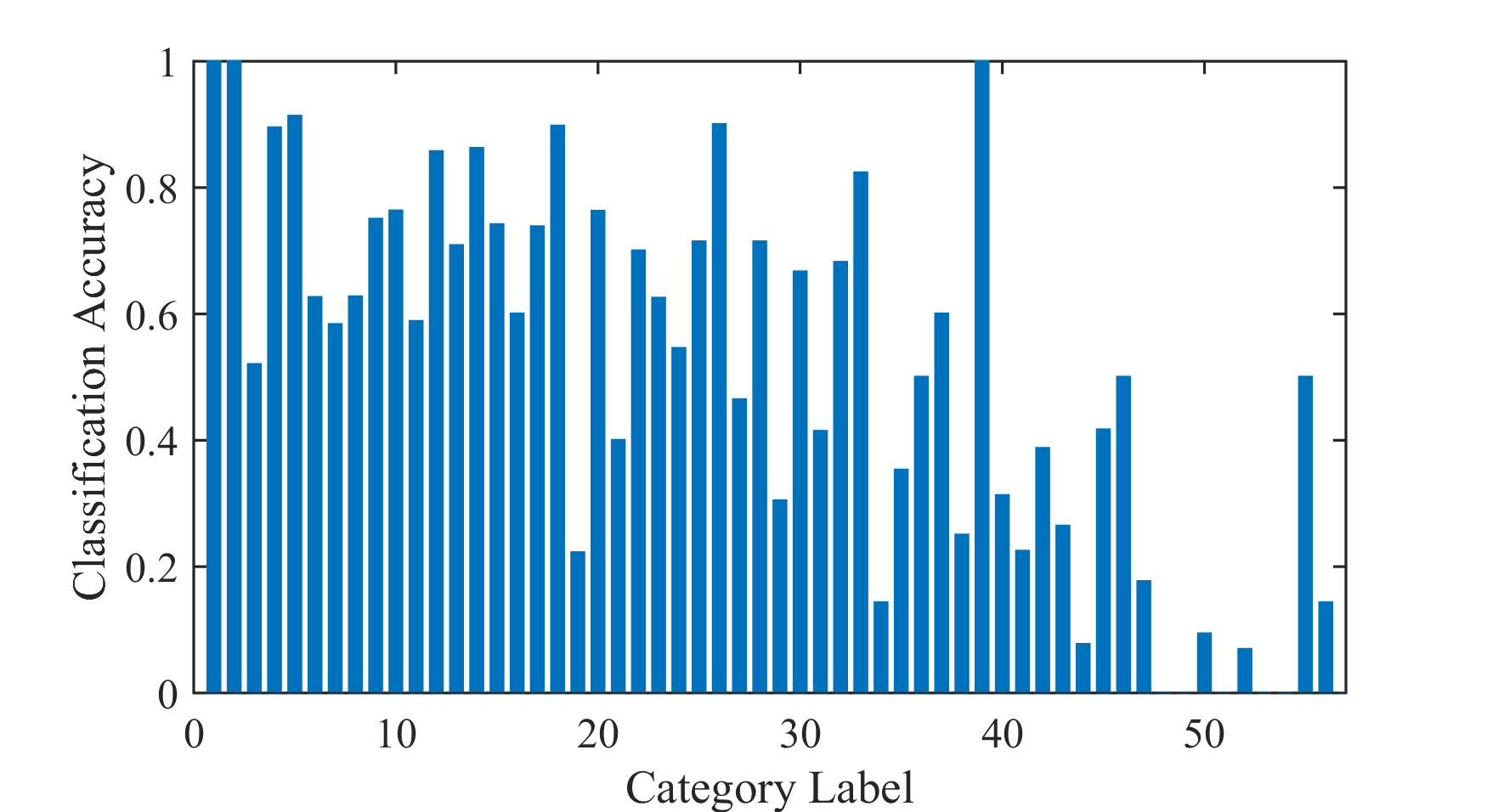}
\caption{Classification accuracy for each class}
\label{fig:accuracy_each_class}
\end{figure}

\begin{figure} [!t]
\centering
\includegraphics[width=\columnwidth]{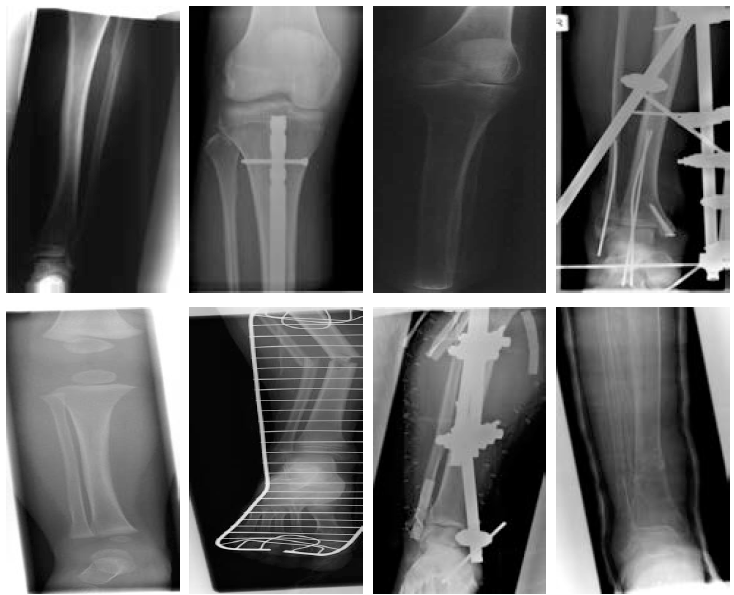}
\caption{Examples from the same class constituting a challenging case for classification and retrieval: The top four images are from the training dataset whereas the bottom four images are from testing set.}
\label{Examples_within_same_category}
\end{figure}

Figures \ref{fig:number_of_images_each_class} and \ref{fig:accuracy_each_class} display the number of images in each class and the classification accuracy for each category, respectively. For some special classes, SVM classifier attains $100\%$ accuracy. However, it can be seen that a high classification accuracy cannot be achieved for every class. Generally, an increase in number of classes may lower the classification accuracy of SVM. Although some images are in the same class, they may still be very difficult to distinguish even for specialists (a consensus may be difficult unless large number of experts evaluate the same images). Figure \ref{Examples_within_same_category} shows an example for such cases (the first four images are from the training dataset whereas the last four images are from testing). It can also be observed that for some specific classes the classification accuracy is zero or close to zero because there are only a few wrongly classified images in that category or even no image of that class is present in the testing data.

\subsection{Image Retrieval Results}
\begin{figure}[!t]
\centering
\includegraphics [width=.5\textwidth]{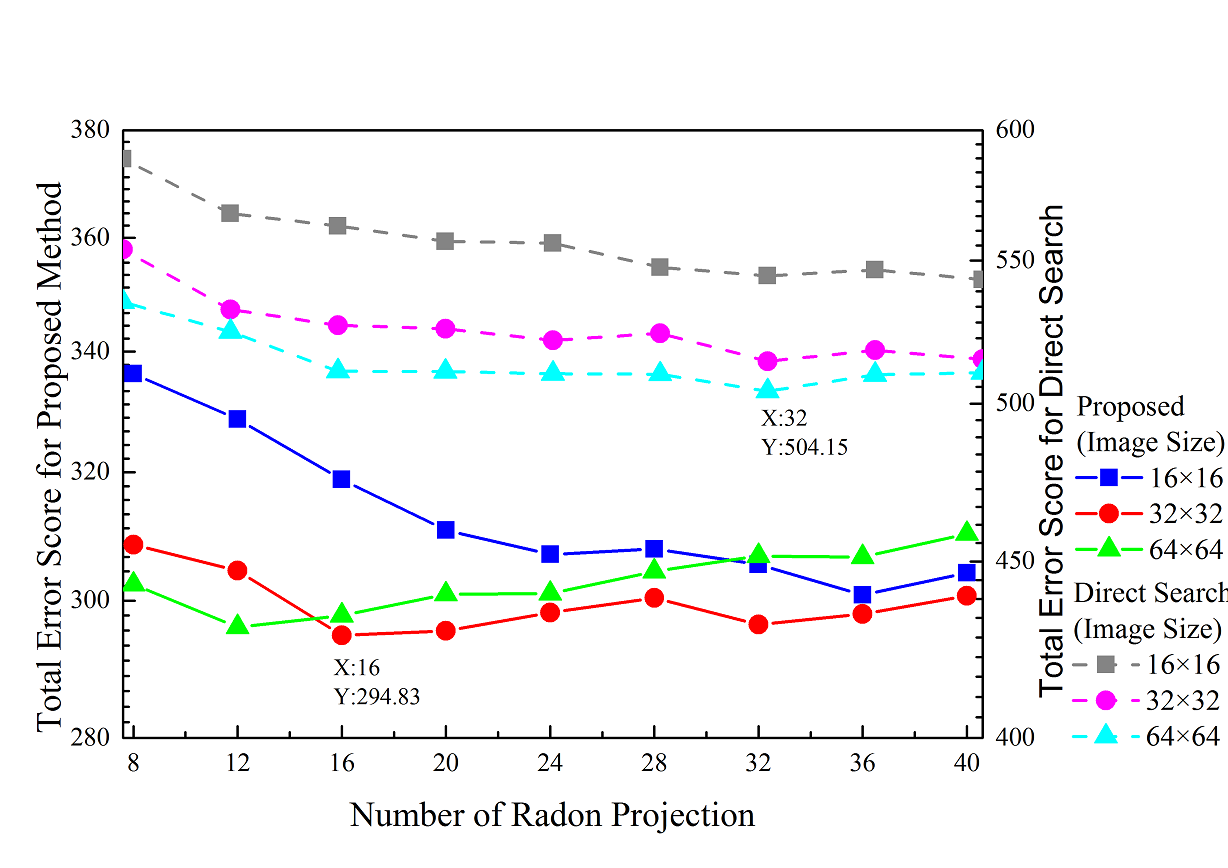}
\caption{Comparison of the proposed method and direct search in total error score}
\label{fig:Total error}
\end{figure}

Figure \ref{fig:Total error} presents the comparison of the direct KNN search and the proposed method in terms of total error score with various normalized image sizes and projection angles. The direct KNN method searches the entire Radon barcode dataset instead of the subset identified by the SVM classifier. It can be observed that, comparing with direct Radon barcode search, the proposed method performs much better and achieves the lowest error at 294.83 From the Figure \ref{fig:Total error}, we can conclude that the performance has a similar trend as Table \ref{tab:SVMresults} indicating that the increase in the number of projection angles generally leads to a better performance; the relatively better performance can be achieved by 32$\times$32 downsampling and 16 projection angles.

Table \ref{tab:compIRMA} provides an overview of reported results in literature \cite{tommasi2010overview}. The proposed approach belongs to methods with lower error rates. Of course, a more comprehensive comparison should include speed, memory and implementation ease. The running time for retrieval stage was 51.5 milliseconds per image in our case. Most papers applied on IRMA dataset do not provide any detail in this regard. 

\begin{table}[!t]
\caption{Error score performance for various methods applied on 2009 IRMA dataset} 
\centering 
\begin{tabular}{lc} 
\hline
Method \& run & Error for IRMA dataset \\ \hline
Camlica et al. \cite{camlica2015SVM} & 146.55 \\
TAUbiomed\_95\_9\_1246120389711 \cite{Avni2011,Avni2009}&169.50 \\ 
Idiap\_3\_9\_1245417716666 & 178.93 \\ 
Idiap\_3\_9\_1245417671272 & 227.19\\
FEITIJS\_96\_9\_1245937057229 & 242.46\\ 
VPA SabanciUniv\_63\_9\_1245419336923 & 261.16\\ 
\textbf{Proposed Method} & \textbf{294.83} \\ 
MedGIFT\_77\_9\_1245961041705 & 317.53\\
VPA SabanciUniv\_63\_9\_1246033855761 & 320.61\\
IRMA & 359.29\\
MedGIFT\_77\_9\_1246044416990 & 420.91\\
VPA SabanciUniv\_63\_9\_1245936277557 & 574.00\\
DEU\_97\_9\_1246226037987 & 642.50\\
DEU\_97\_9\_1245952673253 & 710.10\\
\hline
\end{tabular}
\label{tab:compIRMA}
\end{table}

Although Camlica et al.'s method \cite{camlica2015SVM} rank the first with an error score 146.55 in Table \ref{tab:compIRMA}, their method is rather time-consuming both for training and retrieval procedure. As this method should create a saliency map from input query and extract the multi-scale LBP features from the created saliency image. The generation of this saliency map, based on our experiments with the original code \footnote{http://webee.technion.ac.il/labs/cgm/Computer-GraphicsMultimedia/Software/Saliency/Saliency.html} is quite expensive and takes several seconds for each image. 

The method that ranks the second in Table \ref{tab:compIRMA}, is described in \cite{Avni2009} and \cite{Avni2011}. The method uses many techniques such as dictionary learning, PCA and SVM training, making the algorithm quite complex. It takes approximately 90 minutes for the whole system (training and classification), and 400 milliseconds per query for retrieval on a dual quad-core Intel Xeon 2.33 GHz. Most approaches only report their total error without any specific information about time complexity and memory requirements of their methods. The proposed approach, with a total error of 294.83, ranks in the middle of reported errors for IRMA dataset. However, we have to mention that: 
\begin{enumerate}
\item We only use one technique (Radon transform) to build the SVM classifier and generate the Radon barcode, which makes our method simple and easy to implement; 
\item The memory and storage requirement is very low for our method. For instance, only 80 bytes is needed to store Radon barcode for $32 \times 32$ normalized image with 20 Radon projections (123 bytes in Matlab due to zero-padding of Radon projections). That is 80 MB for one million images and 80 GB for one billion images; 
\item The proposed method only needs 84.49 seconds (parallel computing with 4 workers) for retrieval task for all 1,639 testing images (51.5 milliseconds per image) on a dual quad-core Intel i7 4790 3.60 GHz.
\end{enumerate}

\subsection{Visual Examples}
\begin{figure}
\setlength{\abovecaptionskip}{10pt} 
\setlength{\belowcaptionskip}{-10pt}
\centering
\includegraphics[width=\columnwidth]{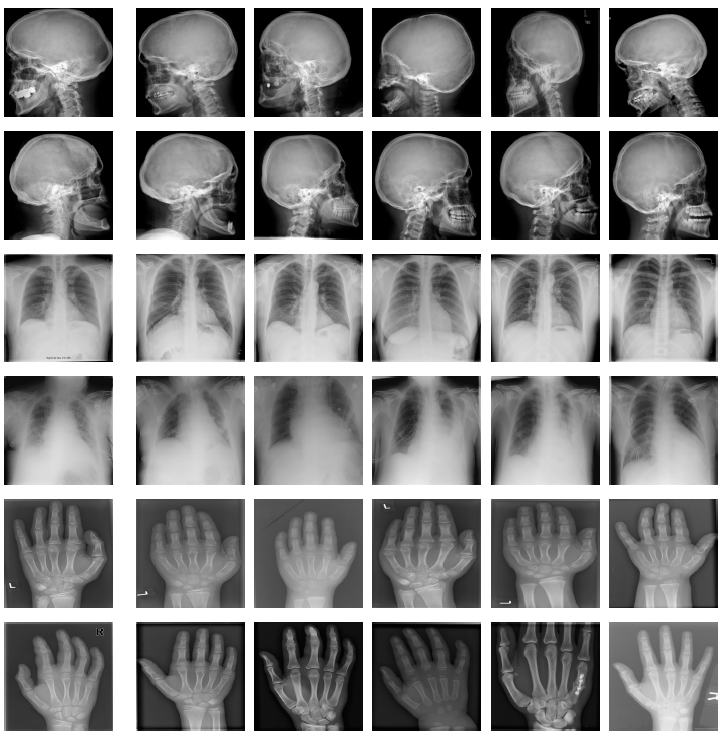}
\caption{Results of query-by-example experiment for correctly identified query}
\label{Fig_correctly_retrieved}
\end{figure}

Some query-by-example experiments are implemented in this part. Figure \ref{Fig_correctly_retrieved} illustrates some samples with correctly classified categories. The images on the left hand side in each row are the original query images and on the right hand side are top 5 retrieved images based on the Hamming distance of the Radon barcodes. The results demonstrate that the proposed method has the ability to retrieve similar images from the same category as long as the class of the query image is identified correctly. 

The query-by-example experiment with regard to the wrongly classified samples by SVM are presented in Figure \ref{Fig_incorrectly_retrieved}. The results show that even if the class of the query images cannot be correctly identified by SVM, the retrieved responses may still be visually similar to the input image. Some of the retrieved images shown in Figure \ref{Fig_incorrectly_retrieved} are supposed to be in the same category as the query ones, which implies that more accurate labeling of the dataset will improve the performance of the proposed approach. Therefore, we can conclude that our method depends on the a performance of the classifier but not limited by it. 

\begin{figure}
\centering
\caption{Results of query-by-example experiment for wrongly identified query}
\label{Fig_incorrectly_retrieved}
\end{figure}
\includegraphics[width=\columnwidth]{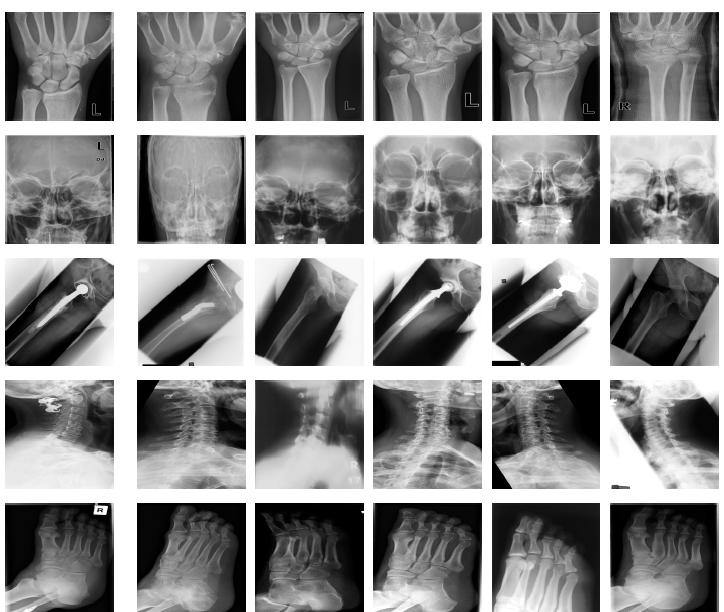}
\section{Conclusion}
\label{sec:conc}
In this work, based on SVM classification and Radon barcodes, a content-based image retrieval method is proposed. The extracted Radon features are used to train a multi-class SVM classifier in order to categorize query images. Radon barcodes which represent the image in a binary format, make it efficient for $k$-nearest neighbors method to search for similar images. Experimental results demonstrate that the proposed method is able to retrieve similar responses to the correctly identified query and even for the incorrectly classified ones. A better classifier or the improved version of current one will be studied in the future work. 

\bibliographystyle{unsrt}
\bibliography{IEEEexample}

\end{document}